\documentclass{article}
\usepackage{arxiv}

\usepackage[utf8]{inputenc} 
\usepackage[T1]{fontenc}    
\usepackage{hyperref}       
\usepackage{url}            
\usepackage{booktabs}       
\usepackage{amsfonts}       
\usepackage{nicefrac}       
\usepackage{microtype}      
\usepackage{lipsum}		
\usepackage{graphicx}
\usepackage{natbib}
\usepackage{doi}

\usepackage{researchpack}
\usepackage[capitalise,nameinlink]{cleveref}
\usepackage{booktabs}
\usepackage{enumitem}
\usepackage{xspace}
\usepackage{multirow}
\usepackage{floatrow}
\usepackage[T1]{fontenc}
\usepackage{graphicx}

\hypersetup{
    colorlinks=true,
    citecolor=blue,
    linkcolor=blue,
    filecolor=blue,
    urlcolor=blue,
}

\newfloatcommand{capbtabbox}{table}[][\FBwidth]

\newcommand{\acronym}{Surrogate-based Learning to Guide\xspace}
\newcommand{\method}{\textsc{slog}\xspace}

\newcommand{\dataset}{\ensuremath{\cal D}\xspace}
\newcommand{\Guidance}{\ensuremath{\cal G}\xspace}
\newcommand{\Score}{\ensuremath{q}\xspace}

\usepackage{color}

\date{}

\title{Learning to Guide Human Experts \\ via Personalized Large Language Models}

\author{Debodeep Banerjee\\
	DI, University of Pisa\\
        DISI, University of Trento\\
	\And
	  Stefano Teso  \\
	CIMeC, university of Trento\\
        DISI, University of Trento\\
	  \AND
	  Andrea Passerini \\
	  DISI, University of Trento \\
}

\renewcommand{\shorttitle}{Learning to Guide}

\begin{document}
\maketitle

\begin{abstract}
    In \textit{learning to defer}, a predictor identifies risky decisions and defers them to a human expert.
    One key issue with this setup is that the expert may end up over-relying on the machine's decisions, due to anchoring bias.  At the same time, whenever the machine chooses the deferral option the expert has to take decisions entirely unassisted.
    As a remedy, we propose \textit{learning to guide} (LTG), an alternative framework in which -- rather than suggesting ready-made decisions -- the machine provides \textit{guidance} useful to guide decision making, and the human is entirely responsible for coming up with a decision.
    We also introduce \method, an LTG implementation that leverages (a small amount of) human supervision to convert a generic large language model into a module capable of generating \textit{textual} guidance, and present preliminary but promising results on a medical diagnosis task.

\end{abstract}

\keywords{
    Hybrid Decision Making \and
    Learning to Defer \and
    Interactive Machine Learning \and
    Large Language Models \and
    Medical Diagnosis.
}

\section{Introduction}
Consider the problem of diagnosing lung pathologies based on the outcome of an X-ray scan.
This task cannot be fully automated, for safety reasons, necessitating human supervision at some step of the process.
At the same time, it is difficult for human experts to tackle it alone due to how sensitive the decision is, especially under time pressure.
High-stakes tasks like this are natural candidates for \textit{hybrid decision making} (HDM) approaches that support human decision makers by leveraging AI technology for the purpose of improving decision \textit{quality} and lowering \textit{cognitive effort}, without compromising \textit{control}.

Most current approaches to HDM rely on a \textit{learning to defer} (LTD) setup, in which a machine learning model first assesses whether a decision can be taken in autonomy -- \ie it is either safe or can be answered with confidence -- and defers it to a human partner whenever this is not the case \citep{madras2018predict, mozannar2020consistent, keswani2022designing, verma2022calibrated, liu2022incorporating}.
Other forms of HDM, like learning to complement \citep{wilder2021learning}, prediction under human assistance \citep{de2020regression}, and algorithmic triage \citep{raghu2019algorithmic, okati2021differentiable} follow a similar pattern.
We argue that this setup is suboptimal and potentially dangerous.  The reason is that there is a risk that humans will end up over-trusting the machine's decisions (a phenomenon known as anchoring bias~\citep{Rastogi2022}).  At the same time, whenever the machine opts for deferral, the human is left resolving hard cases completely unassisted.
Both situations conflict with the aims of HDM.

As a remedy, we propose \textit{learning to guide} (LTG), an alternative algorithmic setup that avoids these issues entirely.
Instead of proposing potential decisions, in LTG the machine supplies its human partner with interpretable \textit{guidance} highlighting aspects of the input $\vx$ that are useful for coming up with a sensible decision.
In this setup, responsibilities cannot be shifted:  by construction, all decisions are taken (under assistance) by the human in the loop.

In this paper, we design an LTG pipeline for medical decision-making that relies on \textit{textual guidance}.
In this context, guidance is implemented as a brief but meaningful description of the symptoms visible in an input X-ray scan, and present \method (\acronym), an approach for converting a generic pre-trained large language model (LLM) into a \textit{hint generation model} tailored to a concrete decision making task.
The conversion is \textit{interactive}:  a human expert provides quality judgments on generated guidance, which \method uses to fine-tune the model itself.  Since human feedback is expensive and therefore scarce, \method makes use of a \textit{surrogate model} that generalizes from a handful of quality judgments to fine-tune the LLM.

\section{Learning to Guide with \method}
\label{sec:method}

\paragraph{Learning to guide for medical diagnosis.}  We consider the problem of diagnosing lung pathologies $y$ from X-ray scans $\vx$.
Rather than learning a classifier for inferring $y$ directly, as in LTD, in LTG the goal is to learn a \textit{guidance generator} that, given $\vx$, extracts \textit{textual guidance} $\Guidance$ in the form of a short caption capturing salient properties of the scan that are useful for supporting human decision making.
Naturally, guidance should be both \textit{interpretable} and \textit{informative}, so that the human decision maker can make a reasonable decision based on it.

\paragraph{The \method algorithm.}  To address this problem, we propose \method.  It requires access to the following elements:
(\textit{i}) A \textit{caption generator} $g: \vx \mapsto (\Guidance, \vz)$, implemented with an LLM, that extracts textual guidance $\Guidance$ as well as the latent representation $\vz$ of $\Guidance$, and
(\textit{ii}) A \textit{decision maker} (DM) that, given $\vx$ and $\Guidance$, comes up with a decision $y$, say {\tt healthy} vs.\@ {\tt pneumonia}, and -- whenever explicitly requested -- also with a \textit{score} $\Score \in \bbR$ summarizing how good the guidance $\Guidance$ is for inferring a decision.

The LLM $g$ is pre-trained and as such it does not generate captions specifically tailored for the decision making task at hand.
The only party capable of determining whether the textual guidance is good enough is the human partner, so in principle, we can use their judgment to fine-tune $g$ so as to produce more useful guidance.
The limiting factor here is annotation cost:  an annotator can only produce so much feedback, making it difficult to fine-tune the LLM with it.

\method tackles these challenges in an iterative fashion.
Let $\dataset = \{\vecvar{x_i} \}$ be a data set of X-ray images.
In each step $t$, \method takes an (initially pre-trained) LLM $g$ and uses it to generate guidance $\Guidance$ and corresponding embeddings $\vz$ for a small set of images $\vx \in \dataset$.
These are shown to the decision maker, who scores all of them.
This way, we obtain a \textit{fine-tuning set} $\calF = \{ (\vx, \vG, \vz, \Score) \}$ exemplifying the human's opinion of generated guidance.
These annotations are then used to fit a \textit{surrogate model} $s: \vz \mapsto \widehat{\Score}$, implemented using an appropriate regression architecture.  The surrogate is responsible for generalizing the (scarce) human feedback and can be used to score \textit{any} guidance generated by the LLM.
Once the surrogate is trained, we freeze it and use it to fine-tune $g$ by minimizing the following augmented loss for a handful of epochs:
\[
    \calL(g, \dataset)
    + \lambda \cdot \bbE_{(\vx, \Guidance, \vz, \ensuremath{q}\xspace) \sim \calF} [
        - \vecvar{s} (\vz)
    ]
\]
%
The first term is the regular LLM loss -- for instance, the negative log-likelihood of the generated text -- on the data set $\dataset$, while the second one is a novel penalty term that encourages the model to generate captions obtaining a high score according to the surrogate.  Here, $\lambda > 0$ is a hyperparameter.
This step increases the overall quality of the generated guidance while making sure that the LLM still outputs sensible captions.
The \method loop then repeats.
Since the LLM's embedding space changes during fine-tuning, the surrogate is fit anew in each iteration.  This operation is very cheap in comparison to fine-tuning the LLM itself.
If the surrogate manages to properly generalize the human's feedback, the LLM gradually learns to output image captions that work well as textual guidance and that are tailored for the specific task and human expert at hand.

\paragraph{Related Work.} Using human guidance to fine-tune large language models has recently become a popular topic.
\citet{Bazi2023, yunxiang2023chatdoctor, wang2023chatcad} proposed medical chat models.  While \citet{Bazi2023} introduced a specially designed vision transformer, the other two opted to fine-tune already available language models. \citet{seo2020reinforcing} presented a method for improving the performance of an image caption generator with offline human feedback. \citet{hou2021ratchet, chen2020generating} focused on machine-driven pathological report generation from chest X-ray images. They experimented with their models using the Mimic-CXR-IV \citep{johnson2019mimic} and Indiana University chest X-ray data \citep{demner2016preparing}.
%
%
None of these approaches are concerned with fine-tuning LLMs for the purpose of generating textual guidance useful for supporting human decision-making.

\section{Empirical Analysis}
\label{sec:experiments}


\paragraph{Data set.}  We evaluate \method on the Mimic-CXR-IV data set \citep{johnson2019mimic}.  The data consists of 377,100 chest X-ray images and 227,827 corresponding radiology reports.
We filtered retained only examples whose reports have information relevant for decision making (specifically, a \textit{findings} or \textit{impression} field). Thereafter, we split the data into train, validation, fine-tuning, and test. In this paper, we use a only subset of this obtained data set for computational ease.
Ground-truth human judgments are derived as follows.  For each of the labels presented in \citet{irvin2019chexpert}, we assign scores of $1$, $-1$, and $0$ depending on the presence, absence, and ambiguous mention or missing information of that particular label in the report.  Then, for each report, we sum over the labels to obtain an aggregate, numerical information score.

\begin{figure}[!t]
    \begin{floatrow}

    \capbtabbox[12em]{%
        
     
        \begin{tabular}{cc}
        \toprule
        {\sc Split} &  {\sc \#Examples}
        \\
        \midrule
        Train & $7,374$  \\	
        Validation & $1,475$  \\		
        Test & $1,627$  \\
        \bottomrule    
    \end{tabular}
    }{%
        \caption{Train, validation, and test split for training the surrogate model}%
        \label{tab:data}
    }
    \ffigbox{%
        \centering
        \includegraphics[height=7.8em]{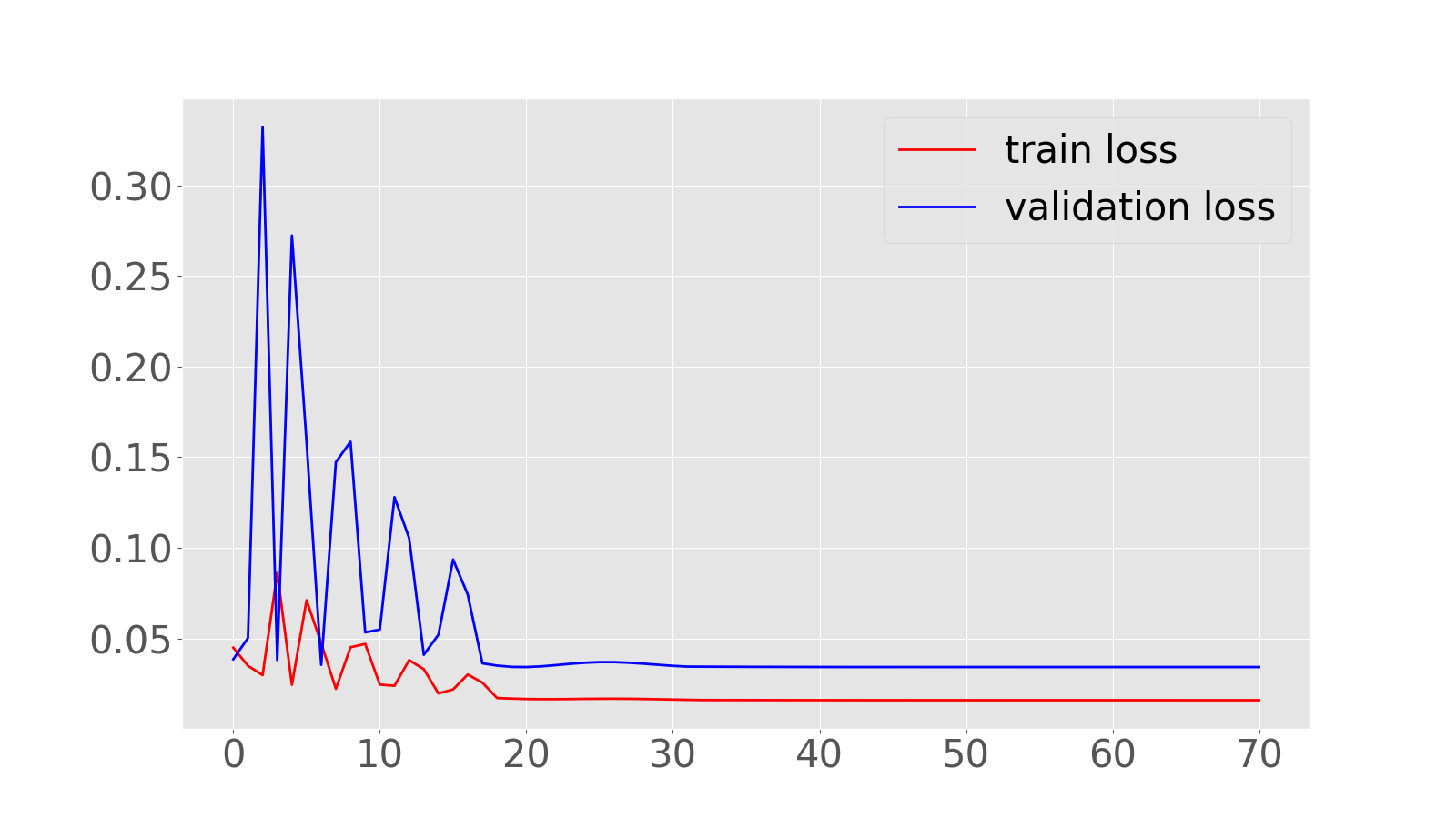}
    }{%
        \caption{Training and validation loss of our non-linear ridge regression surrogate model.}%
        \label{fig:surrogate}
    }

    \end{floatrow}
\end{figure}

\paragraph{Architectures and metrics.}  We used the offline LLM architecture developed by \cite{chen2020generating}. for generating pathological reports from chest X-ray images.
The LLM is a memory-driven transformer with a relational memory layer recording additional information and then used during decoding.
%
The surrogate model itself is a non-linear ridge regression model.
Since the latent representation $\vz$ of the reports in the training set is not available, we generate it using the pre-trained LLM and weight each example using the BLEU4 score \citep{papineni2002bleu} of the corresponding generated text. The surrogate model is then fit on this data using a weighted loss.

\paragraph{Results.}  The RMSE on the training and test sets are reported in \cref{fig:surrogate}.  We observe that the surrogate model quickly fits the training examples (in \textbf{\textcolor{red}{red}}) while performing well on the validation data (in \textbf{\textcolor{blue}{blue}}). The split of the data is reported in \cref{tab:data}  When tested on a separate set of $1627$ test inputs, the test RMSE is $0.1828$, which is well within the variance of the ground-truth human judgments ($0.1817$).
This allows us to conclude that even simple non-linear models can in fact be used to generalize human judgments of textual guidance in this setting.

\section{Conclusion}
\label{sec:conclusion}

We introduced \textit{learning to guide} as an alternative setup for hybrid decision making that ensures the human is always in the loop, as well as \method, an end-to-end approach for fine-tuning an LLM to produce high-quality textual guidance.  Our preliminary results suggest that it is in fact possible to generalize human judgments using a surrogate model, supporting the feasibility of our approach.  In future work, we plan to properly evaluate the quality of textual guidance that can be obtained using \method.

\paragraph{Acknowledgments.}  The research of ST and AP was partially supported by TAILOR, a project funded by EU Horizon 2020 research and innovation programme under GA No 952215.







\bibliographystyle{unsrtnat}
\bibliography{main}
\end{document}